\documentclass{article}
\usepackage{ijcai26}

\usepackage{times}
\usepackage{soul}
\usepackage{url}
\usepackage[hidelinks]{hyperref}
\usepackage[utf8]{inputenc}
\usepackage[small]{caption}
\usepackage{graphicx}
\usepackage{amsmath}
\usepackage{amsthm}
\usepackage{booktabs}
\usepackage[ruled,vlined]{algorithm2e}
\usepackage{subcaption}
\usepackage[switch]{lineno}
\usepackage[table]{xcolor}
\usepackage{multirow}
\usepackage{booktabs, multirow, tabularx, adjustbox}
\usepackage{colortbl}
\usepackage{diagbox}

\title{FedUP: One-Shot Federated Unlearning via Centroid-Guided Plug-in Filters}

\author{
Feihong Nan$^{1}$\and
Zhengyi Zhong$^{1}$\and 
Pan Wang$^{1}$\and 
Weidong Bao$^{1}$\and 
Xiongtao Zhang$^{1}$\and \\
Quan Wen$^{1}$\And 
Ji Wang$^{1}$\thanks{Corresponding Author}\\ 
\affiliations
$^1$National Key Laboratory of Big Data and Decision, National University of Defense Technology, China.\\
\emails
feihongnan178@outlook.com,
\{zhongzhengyi20, wangpan19, wdbao, zhangxiongtao14, wangji\}@nudt.edu.cn,
weixingw1@sina.com
}

\begin{document}

\maketitle

\begin{abstract}
    Federated unlearning (FU) is critical for complying with legal mandates like the right to be forgotten in decentralized systems, yet current methods face a persistent dilemma between \textit{non-target knowledge loss} and \textit{high request latency}. To resolve these issues, we propose FedUP, a one-shot federated unlearning framework utilizing lightweight pluggable filters that act as a ``knowledge funnel" to screen out target data while preserving original model performance. By freezing original model parameters and training filters at the server side using differentially private (DP)-protected class centroid samples, FedUP bypasses the need for multi-round client-server communication and complex retraining, reducing unlearning latency from minutes to mere seconds. Additionally, the framework's pluggable architecture ensures inherent reversibility, enabling the seamless restoration of forgotten knowledge by simply removing the filters. Extensive experiments on diverse image and text tasks demonstrate that FedUP effectively reduces non-target knowledge loss and achieves superior unlearning precision and efficiency across various scenarios. Code is available at: \url{https://github.com/suows/FedUP-code}.
\end{abstract}
  
\section{Introduction}
\textbf{Background}. As a distributed machine learning paradigm, federated learning (FL) \cite{McMahan_Moore_Ramage_Hampson_Arcas_2016,truong2021privacy,zhong2022flee,qi2024cross,zhong2025sacfl,fu2025learn,jiang2026unveiling} has gained significant attention in privacy-sensitive scenarios recently because it eliminates the need for centralizing raw data during training. However, during the training process, the global model internalizes client information into its parameters through multiple rounds of parameter aggregation, which exposes novel privacy risks when the model is confronted with legal mandates such as the right to be forgotten under GDPR \cite{de_la_Torre_2018}. To address this, researchers have focused on Federated Unlearning (FU), aiming to remove specific knowledge without retraining the global model from scratch. Existing FU methods generally follow two main technical paradigms: server-side FU methods \cite{wu2022federated,huynh2025certified,pan2025federated}, which implement approximate unlearning \cite{yang2025erase} on the global model at the server side; and client-side FU methods \cite{wang2023bfu,liu2022right,zhu2023heterogeneous,zhong2025unlearning}, which achieve exact unlearning \cite{kuo2025exact} through iterative client-side model retraining procedures.

\begin{figure}[t]
    \centering
    \begin{subfigure}{0.485\linewidth}
        \includegraphics[width=\linewidth]{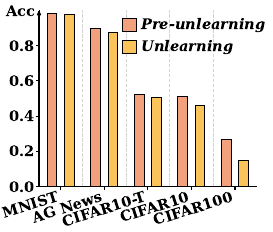}
        \caption{Non-target knowledge loss of server-side FU method.}
        \label{fig:Non-target Knowledge Loss}
    \end{subfigure}
    \begin{subfigure}{0.485\linewidth}
        \includegraphics[width=\linewidth]{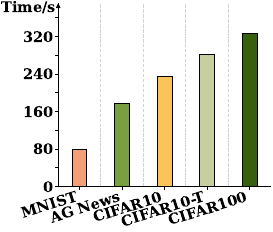}
        \caption{Unlearning request latency of client-side FU method.}
        \label{fig:response_time_retain_bar}
    \end{subfigure}
    \caption{FedEraser, a representative server-side method, shows noticeable non-target knowledge loss after unlearning across multiple tasks. Client-side methods like FUSED experience significantly longer convergence times as task difficulty increases, with responses up to 5 minutes indicating severe unlearning request latency.}
    \label{fig:challenges}
\end{figure}

\textbf{Existing Challenges}. While current Federated Unlearning (FU) methods facilitate knowledge removal to various extents, challenges regarding non-target knowledge loss and unlearning request latency still remain. As depicted in Figure \ref{fig:challenges}, server-side FU methods lack precise control over the unlearning scope, frequently incurring \textbf{\textit{non-target knowledge loss}}, where the model performance unrelated to the unlearning request is inevitably impaired during knowledge removal. Conversely, although client-side FU methods achieve exact unlearning via retraining-based approaches, they are heavily constrained by multi-round training and communication, leading to \textbf{\textit{high request latency}}. To date, it is hard to find a solution that simultaneously ensures rapid response and prevents non-target knowledge loss. Moreover, both server-side and client-side FU paradigms typically rely on direct modification of original model parameters, which leads to \textbf{\textit{the irreversibility of unlearning}}. Once the unlearning process is finalized, restoring previously removed knowledge necessitates further parameter tuning or retraining, thereby imposing complexity and additional overhead on practical deployment.

\begin{figure}[htbp]
    \centering
    \includegraphics[width=0.97\linewidth]{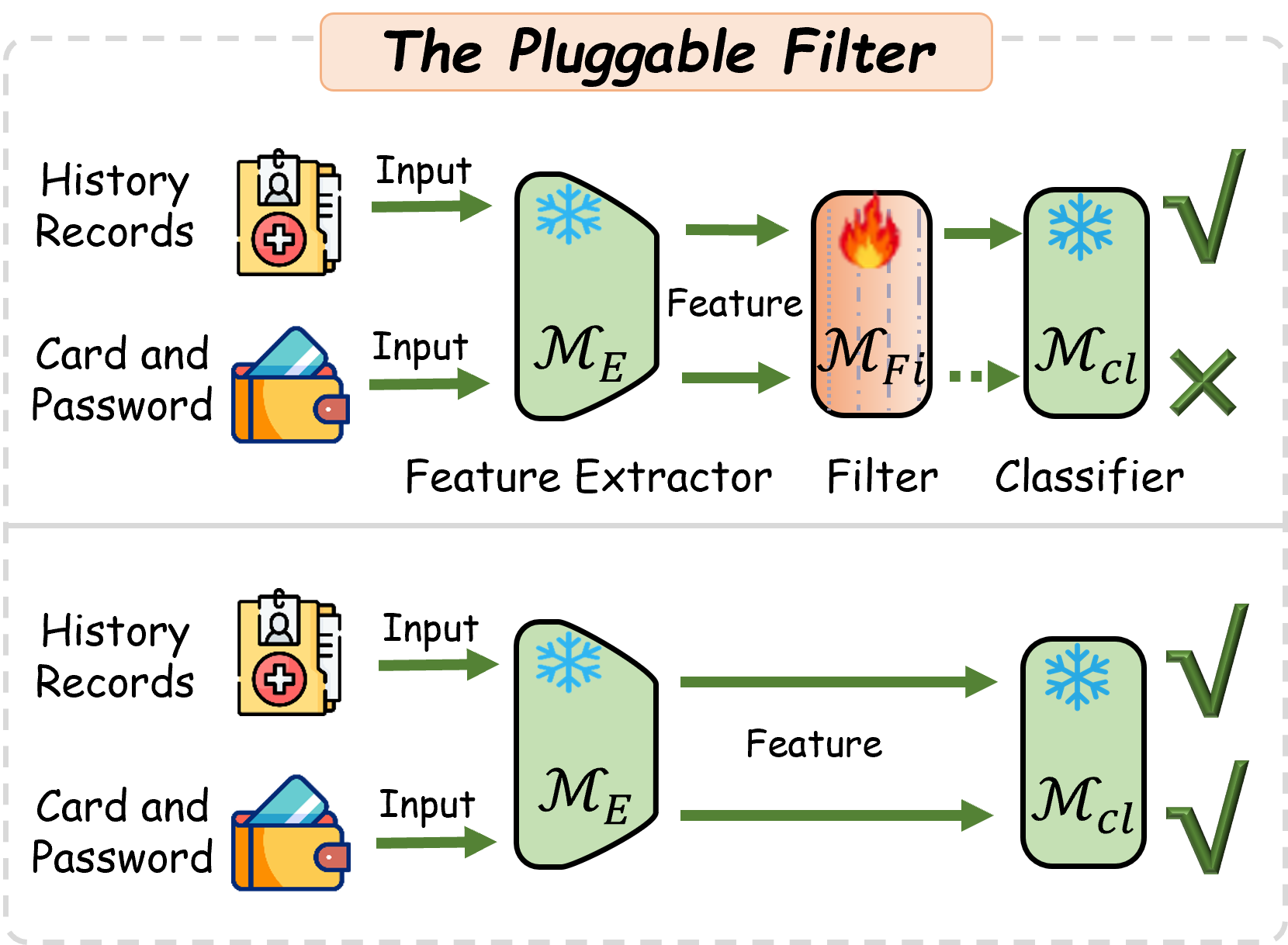}
    \caption{Lightweight plug-in filters.}
    \label{filter}
\end{figure}

\textbf{Proposed Solution}. To this end, we propose FedUP, a one-shot federated unlearning framework based on lightweight pluggable filters. In terms of mitigating non-target knowledge loss, the framework avoids performing direct, large-scale parameter updates on the global model; instead, it implements unlearning by introducing independent pluggable filters while keeping original model parameters frozen (shown in Figure \ref{filter}). These filters serve as a ``knowledge funnel" that screens out target knowledge and permits only non-target knowledge to pass, thereby reducing the interference of unlearning on non-target knowledge. To lower unlearning request latency, FedUP only needs to perform a few rounds of fine-tuning on the lightweight filters at the server side using differential privacy (DP)-protected class centroid samples to complete the unlearning task. This bypasses multi-round retraining and frequent communication with clients, significantly accelerating the response speed. Moreover, since the filters are pluggable, the framework inherently supports reversibility. As depicted in Figure \ref{filter}, when restoration of forgotten knowledge is required, it can be achieved simply by removing the filters. Overall, FedUP provides a solution that balances precision, efficiency, and recoverability.

\textbf{Contributions}. The main contributions are as follows: 
\begin{itemize} 
    \item We design FedUP, a one-shot FU framework utilizing DP-protected class centroids, mitigating non-target knowledge loss and reducing unlearning request latency from minutes to seconds.
    \item We propose a reversible unlearning mechanism via lightweight pluggable filters without altering original model parameters while ensuring rapid transitions between unlearned and pre-unlearning states.
    \item We conduct differential privacy analysis showing that appropriate noise protects class centroid samples while minimally impacting data utility. Extensive experiments on image and text tasks confirm FedUP's effectiveness across various scenarios.
\end{itemize}

\section{Related Work}\label{Related Work} 

\textbf{Machine Unlearning}. 
Simply removing training data from storage fails to purge its influence on deployed models. Machine Unlearning (MU) \cite{ma2022learn} is thus introduced to erase such learned knowledge.
Existing methods are categorized as either exact \cite{Yinzhi2018Efficient} or approximate unlearning \cite{golatkar2020eternal}.
Exact unlearning mandates that the post-forgetting model be statistically indistinguishable from one retrained from scratch without the deleted data. It retains the guarantees of complete retraining but lowers its cost via algorithmic shortcuts.
For classical models, the training process can be expressed in an invertible additive form, enabling point removal by subtracting its closed-form term \cite{Yinzhi2018Efficient}.
For complex architectures, exact unlearning employs data sharding \cite{bourtoule2021machine}, localized retraining \cite{chen2022graph}, or intermediate checkpointing \cite{wang2023inductive} to reduce computation while preserving retraining-level guarantees.
Approximate unlearning, in contrast, substitutes full retraining with lightweight fine-tuning, permitting a bounded, residual influence from the deleted data. Related work is typically grouped into data-driven and model-driven approaches \cite{nguyen2025survey}. Data-driven methods involve relabeling retained samples \cite{graves2021amnesiac} or partitioning data shards \cite{gupta2021adaptive} before fine-tuning. Model-driven approaches adjust parameters directly via influence functions \cite{guo2019certified}, fisher-based regularization \cite{golatkar2020eternal}, or knowledge distillation \cite{kurmanji2023towards}, countering the gradient contributions of the data to be erased.

\textbf{Federated Unlearning.}
Though federated learning maintains client privacy through local retention, knowledge from distributed datasets persists in the aggregated global model. This requires integrating MU techniques into FL, named federated unlearning (FU) \cite{wang2022federated}. Based on the execution location, FU methods can be categorized into server-side and client-side methods \cite{liu2024survey,li2025machine}. Server-side methods are intrinsically based on approximate unlearning to expunge client contributions without client involvement. FedEraser \cite{liu2021federaser} calibrates update trajectories but still requires auxiliary retraining. Conversely, Wu et al. \cite{wu2022federated} bypass client-side computation by subtracting historical updates and employing knowledge distillation. To optimize efficiency, Huynh et al. \cite{huynh2025certified} utilize selective retention of influential updates to reduce memory overhead, while Pan et al. \cite{pan2025federated} resolve parameter conflicts via Orthogonal Steepest Descent to accelerate the unlearning process.
Client-side FU methods perform unlearning locally, striving to achieve exact unlearning while balancing computational efficiency with global model utility. Mora et al. \cite{mora2024fedunran} propose FedUNRAN, utilizing local random label perturbations to disperse target gradients and attenuate their influence without server-side intervention. Wang et al. \cite{wang2023bfu} employ variational Bayesian inference for parameter self-sharing to erase target data while preserving performance. To accelerate unlearning, Liu et al. \cite{liu2022right} approximate the Hessian via a diagonal empirical Fisher Information Matrix for quasi-Newton optimization, while Zhu et al. \cite{zhu2023heterogeneous} combine inverse perturbation with passive decay, propagating updates via knowledge distillation. Furthermore, Deng et al. \cite{deng2024enable} introduce model contrastive unlearning (MCU) to regularize the feature space.  Zhong et al. \cite{zhong2025unlearning} implement reversible unlearning through local fine-tuning.

In summary, existing unlearning approaches typically involve trade-offs between unlearning precision and computational efficiency, and often suffer from limited generalization. Moreover, reversibility is rarely considered in current studies. Achieving a unified balance among unlearning precision, efficiency, and reversibility therefore remains an open challenge.

\section{Methodology}\label{Methodology} 

\subsection{Problem Description}\label{RW} 
In the FL framework, a set of clients denoted as $\mathcal{C} = \{C_1, C_2, \ldots, C_N\}$ collaboratively train a global model $\mathcal{M}_G$ without sharing local data. Each client $C_n$ trains a local model $\mathcal{M}_n$ using its local dataset $\mathcal{D}_n$ and uploads the model parameters to a server. The server aggregates these local models via weighted averaging based on dataset sizes to obtain a global model $\mathcal{M}_G$. This process iterates over $W$ global rounds, where each global round comprises $e$ local training epochs on the clients with a local learning rate $l_c$.
$\mathcal{M}_G$ is structurally decomposed into a feature extractor $\mathcal{M}_E$ and a classifier $\mathcal{M}_{cl}$. After aggregation, $\mathcal{M}_G$ is distributed back to all clients, where the feature extractor $\mathcal{M}_E$ is leveraged to extract features from local data. Additionally, a pluggable filter $\mathcal{M}_{Fi}$ is constructed to implement the filtering of knowledge that needs to be forgotten.
The overall dataset is denoted as $\mathcal{D} = \{\mathcal{D}_n\}_{n=1}^{N}$, where $\mathcal{D}_n$ represents the local dataset of client $C_n$. The total data volume across the federation is $\| \mathcal{D} \|= \sum_{n=1}^{N} \| \mathcal{D}_n \|$.
To facilitate data management, particularly for future unlearning operations, each local dataset $\mathcal{D}_n$ is partitioned into two disjoint subsets: a retained subset $\mathcal{D}_n^R$ for model training, and an unlearning subset $\mathcal{D}_n^U$ designated to be removed in compliance with privacy or regulatory constraints.
Let  $k \in \{1, 2, \ldots, K\}$  index the data categories, where  $K$  is the total number of classes.

In the FL phase, the training objective is defined as:
\begin{equation}
    \min_{\theta_{\mathcal{M}_G}} F(\theta_{\mathcal{M}_G}) = 
    \sum_{n=1}^{N} \frac{|\mathcal{D}_n|}{|\mathcal{D}|} 
    \sum_{(x_i^k, y_i^k) \in \mathcal{D}_n} 
    \mathcal{L}\bigl(f(x_i^k; \theta_{\mathcal{M}_G}), y_i^k\bigr),
\end{equation}
where $\mathcal{L}$ denotes the loss function. 
During the FU phase, the objective is to maximize the loss on the unlearning set, minimize the loss on the retained set, and keep the training cost low, described as:
\begin{equation}
    \min_{\theta_{\mathcal{M}_G'}} F^R(\theta_{\mathcal{M}_G'}) = 
    \sum_{n=1}^{N} \frac{|\mathcal{D}_n^R|}{|\mathcal{D}^R|} 
    \sum_{(x_i^k, y_i^k) \in \mathcal{D}_n^R} 
    \mathcal{L}\bigl(f(x_i^k; \theta_{\mathcal{M}_G'}), y_i^k\bigr),
\end{equation}

\begin{equation}
    \max_{\theta_{\mathcal{M}_G'}} F^U(\theta_{\mathcal{M}_G'}) = 
    \sum_{n=1}^{N} \frac{|\mathcal{D}_n^U|}{|\mathcal{D}^U|} 
    \sum_{(x_i^k, y_i^k) \in \mathcal{D}_n^U} 
    \mathcal{L}\bigl(f(x_i^k; \theta_{\mathcal{M}_G'}), y_i^k\bigr),
\end{equation}

\begin{equation}
    \min F^T(\theta_{\mathcal{M}_G'}) = \sum_{w=1}^{W} \text{Comm}^{(w)}(\mathcal{M}_G, \mathcal{C}\bigr),
\end{equation}
where Comm represents the communication resource between the server and the clients $\mathcal{C}$ during each round.

\subsection{Method Overview}

\begin{figure}[!h]
    \centering
    \includegraphics[width=0.475\textwidth]{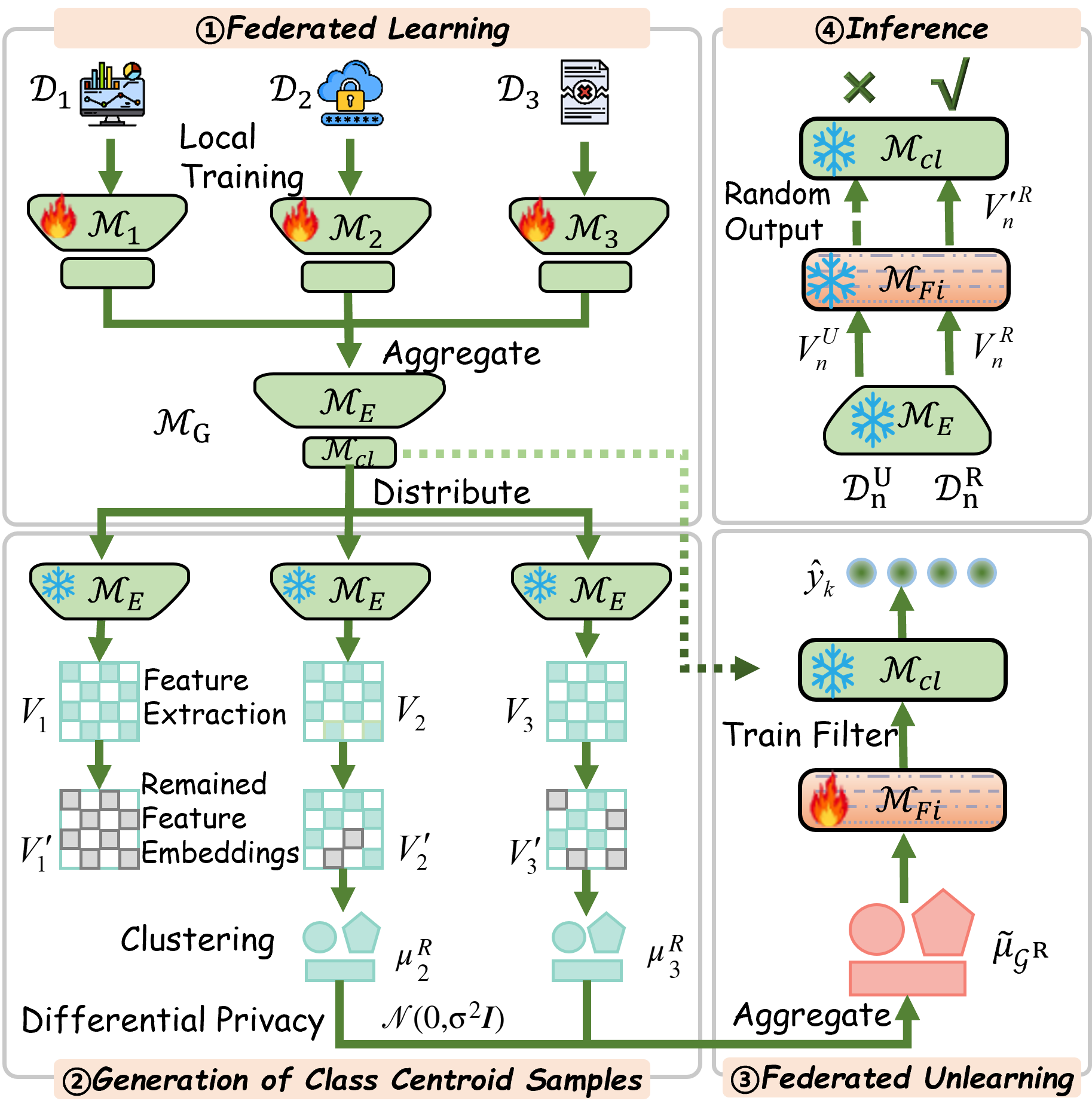}
    \caption{FedUP follows a structured workflow: it begins with federated learning to obtain a global model. Upon request, clients produce differentially‑private class centroid samples from retained data and upload them for server aggregation. The server then fine‑tunes a pluggable filter, blocking forgotten knowledge without modifying the base model. }
    \label{Framework}
\end{figure}

As shown in Figure \ref{Framework}, our method comprises four phases: federated learning, generation of class centroid samples, one-shot federated unlearning, and inference, with each stage's main procedures and key formulas shown in the diagram.

\subsubsection{Feature Extraction}
 Each client $C_n$ performs $e$ rounds of training on its local dataset $\mathcal{D}_n$, resulting in an updated local model $\mathcal{M}_n^{w,e}$. The local training process can be expressed as: 
\begin{equation} 
    \theta_{\mathcal{M}_n}^{w,e} = \theta_{\mathcal{M}_n}^{w,e-1} - l_c \cdot \nabla_{\theta} \mathcal{L}(\mathcal{M}_n^{w,e-1}, \mathcal{D}_n),
\end{equation} 
where $\mathcal{L}$ represents the local loss function and $\nabla_{\theta} \mathcal{L}$ denotes its gradient with respect to the model parameters.Subsequently, the server collects the local models $\mathcal{M}_n^{w,e}$ from selected clients $C_n$ and updates the global model through a weighted average: 
\begin{equation} 
    \theta_{\mathcal{M}_G}^{w+1} = \sum_{n=1}^{N} \frac{|\mathcal{D}_n|}{|\mathcal{D}|} \theta_{\mathcal{M}_n}^{w,e}.
\end{equation} 
The trained global model \( \mathcal{M}_G \) is regarded as a combination of a feature extractor \( \mathcal{M}_E \) and a classifier \( \mathcal{M}_{cl} \). After freezing this composite model, it is deployed to the clients. The features of all data $\mathcal{D}_n$ from each client $C_n$ are extracted by the feature extractor $\mathcal{M}_E$, which can be represented as follows:
\begin{equation} 
    V_n^k = \left\{ \mathcal{M}_E(x_i^k) \mid x_i^k \in \mathcal{D}_n^k \right\},
\end{equation} 
where $V_n^k$ denotes the feature embedding of the data belonging to class $k$ on client $C_n$.

\subsubsection{Generation of Class Centroid Samples}
Upon receiving an unlearning request, client $C_n$ removes the features associated with the unlearning data. The remaining feature embeddings belonging to class $k$ are denoted as:
\begin{equation}
    V_n^{R,k} = V_n^k \setminus \left\{ \mathcal{M}_E(x_i) \mid x_i \in \mathcal{D}_n^{U,k} \right\}.
\end{equation}

To reduce communication cost, the retained features are compressed via KMeans clustering with
\begin{equation}
    K_{n,k} = \lceil \rho \, |V_n^{R,k}| \rceil,
\end{equation}
the number of clusters is set proportionally according to the scenario by $\rho$, yielding a compact set of centroids:
\begin{equation}
    \mu_n^{R,k} = \mathrm{KMeans}(V_n^{R,k}, K_{n,k}).
\end{equation}

For privacy preservation, add noise to each centroid:
\begin{equation}\label{noise}
\tilde{\mu}_{n}^{R,k}
=
\left\{
\mu + \mathbf{z}
\;\middle|\;
\mu \in \mu_{n}^{R,k},
\;
\mathbf{z} \sim \mathcal{N}(0,\sigma^{2}I)
\right\}.
\end{equation}
Here, $\mathcal{N}(0, \sigma^2 \mathbf{I})$ represents a multivariate Gaussian distribution. These mechanisms ensure that the release of $\tilde{\mu}_n^{R,k}$ does not reveal excessive information about any individual data point. Client $C_n$ then uploads all perturbed class centroids $\tilde{\mu}_n^{R,k}$ to the server.
Using the local class centroids $\tilde{\mu}_n^{R,k}$, the server aggregates them to generate the global class centroids $\tilde{\mu}_\mathcal{G}^{R,k}$. The global class centroid is denoted as:
\begin{equation}
    	\tilde{\mu}_\mathcal{G}^{R,k}= \left[	\tilde{\mu}_1^{R,k}; 	\tilde{\mu}_2^{R,k}; \ldots; 	\tilde{\mu}_N^{R,k}\right].
\end{equation}

\subsubsection{One-shot Federated Unlearning}
The objective of the FU phase is to train the filter $\mathcal{M}_{Fi}$ such that it effectively blocks the flow of unlearning knowledge while allowing the retained learning knowledge to pass. The specific steps and formulas are as follows.

A filter $\mathcal{M}_{Fi}$ is inserted into the global model $\mathcal{M}_G=\mathcal{M}_E \oplus \mathcal{M}_{cl}$, yielding a new model structure:
\begin{equation} 
    \mathcal{M}_G' = \mathcal{M}_E \oplus \mathcal{M}_{Fi} \oplus \mathcal{M}_{cl}.
\end{equation}

Subsequently, the parameters of the global model $\mathcal{M}_{G}$ including both $\mathcal{M}_{E}$ and $\mathcal{M}_{cl}$ are frozen:
\begin{equation} 
     \theta_{\mathcal{M_G}} = \theta_{\mathcal{M_G}}^{*}.
\end{equation}

The filter $\mathcal{M}_{Fi}$ is trained using the global class centroids $\tilde{\mu}_\mathcal{G}^{R,k}$. The filtered class centroid prediction is defined as:
\begin{equation}
    \{\hat{y_k}\} = \{\mathcal{M}_{\text{cl}}(\mathcal{M}_{\text{Fi}}(\tilde{\mu})) \mid 	\tilde{\mu} \in \tilde{\mu}_\mathcal{G}^{R,k}, k \in \mathcal{K}_R\}.
\end{equation}

We propose a composite loss function for the filter consisting of cross-entropy and reconstruction losses. 
It preserves discriminative capability on non-target knowledge while enforcing structural consistency in the feature space, thereby improving the stability of the unlearning process without compromising knowledge selectivity.
The cross-entropy loss, measuring the difference between predicted class probabilities and true labels, is defined as:
\begin{equation} 
\mathcal{L}_{\text{CE}} =-\sum_{k} y_k \log(\hat{y}_k),
\end{equation}
where \( y_k \) represents the true label while \( \hat{y}_k \) denotes the corresponding predicted probability for class \( k \).
The reconstruction loss, which measures how well the filter reconstructs its input, is defined as:
\begin{equation} 
\mathcal{L}_{\text{RE}} = \frac{1}{d} \sum_{i=1}^{d} {|\tilde{\mu} - \mathcal{M}_{Fi}(\tilde{\mu})|^2 },
\end{equation}
where \( d \) is the input and output dimensionality of the filter.
The total loss function used to train the filter $\mathcal{M}_{Fi}$ is a weighted sum of these two losses:
\begin{equation}\label{filter loss}
    \mathcal{L}_{\text{total}} = \alpha \mathcal{L}_{\text{CE}} + (1 - \alpha) \mathcal{L}_{\text{RE}},
\end{equation}
where $\alpha$ is a weighting factor that balances the importance of the cross-entropy loss and the reconstruction loss. Train the filter for $W_a$ rounds with a learning rate $l_a$. The update process is as follows:
\begin{equation}\label{update parameters}
    \theta_{\mathcal{M}_{Fi}}^w = \theta_{\mathcal{M}_{Fi}}^{w-1} - {l_a} \cdot \nabla_{\theta} \mathcal{L}_{\text{total}}\left( \theta_{\mathcal{M}_{Fi}}^{w-1}; \tilde{\mu}_\mathcal{G}^{R,k} \right)
\quad w = 1, \dots, W_a.
\end{equation}

When a client requests the restoration of forgotten knowledge, the filter $\mathcal{M}_{Fi}$ can be removed, thereby reverting to the original global model $\mathcal{M}_G$:
\begin{equation} 
    \mathcal{M}_G = \mathcal{M}_E \oplus \mathcal{M}_{cl}. 
\end{equation}

\subsubsection{Differential Privacy Guarantee}
Our framework injects Gaussian noise for privacy guarantee during client-side centroid uploads, satisfying ($\varepsilon,\delta$)-DP. The key is to calibrate the noise scale $\sigma$ according to the desired privacy budget and the $\ell_2$-sensitivity, which is bounded by: 
\begin{equation}
    \Delta_2 f_i = \max_{p,q} \frac{ \| d_p^{(i)} - d_q^{(i)} \|_{2} }{n_i}. 
\end{equation}

\begin{table}[h]
\centering
\fontsize{8pt}{9pt}\selectfont
\setlength{\arrayrulewidth}{1pt} 
\resizebox{\linewidth}{!}{
\setlength{\tabcolsep}{9pt}
\renewcommand{\arraystretch}{1.18}
\newcolumntype{l}{>{\centering\arraybackslash}p{1.99cm}} 
\begin{tabular}{l | c c c c c}
\hline

\rowcolor{gray!20}
\multicolumn{1}{c|}{} 
& \multicolumn{5}{c}{\textbf{\textit{$\sigma$}}} \\

\rowcolor{gray!20}
\multicolumn{1}{c|}{\multirow{-2}{*}{\diagbox[width=2.6cm,font=\bfseries\itshape]{\hspace{-15mm}Dataset}{Noise}}}
& \textbf{0.001} & \textbf{0.005} & \textbf{0.01} & \textbf{0.05} & \textbf{0.1} \\
\hline
\textbf{\textit{MNIST}} 
& \textbf{11.09} & 2.22 & 1.11 & 0.22 & 0.11 \\
\textbf{\textit{CIFAR-10}} 
& \textbf{16.73} & 3.35 & 1.67 & 0.33 & 0.17 \\
\textbf{\textit{AG News}} 
& \textbf{12.27} & 2.45 & 1.23 & 0.25 & 0.12 \\
\textbf{\textit{CIFAR-100}} 
& 141.40 & 28.28 & \textbf{14.14} & 2.83 & 1.41 \\

\hline
\end{tabular}
}
\caption{Average privacy budget for different datasets.}
\label{tab:privacy budget}
\end{table}

We set $\sigma \geq \frac{\sqrt{2\ln(1.25/\delta_i)} \cdot \Delta_2 f_i}{\varepsilon_i}$ with $\delta_i = 1/n_i$. Empirical $\varepsilon_i$ values (Table \ref{tab:privacy budget}) confirm moderate privacy budgets $\varepsilon \approx 10$ \cite{wei2020federated} for all datasets under specified $\sigma$. 
We calculate our differentially private guarantee as:

\begin{equation}
	\varepsilon_{i} = \frac{\sqrt{2 \ln(1.25 n_{i})} \cdot \Delta_{2} f_{i}}{ \sigma}.
\end{equation}
The final choice of the optimal $\sigma$ is determined via $\sigma = q \cdot s$, with the corresponding values of $q$ and $s$ established accordingly. Detailed analysis and supporting experiments are relegated to Section 2.1 and 2.2 of the supplementary material.

\subsection{Algorithm}

\begin{algorithm}[t]
    \caption{FedUP}
    \label{alg:one-shot}
    \textbf{Input}: Number of global rounds $W$, local rounds $e$, local learning rate $l_c$, adapter learning rate $l_a$, clients $\mathcal{C} = \{C_1, C_2, \ldots, C_N\}$, dataset $\mathcal{D} = \{\mathcal{D}_n\}_{n=1}^N$\\
    \textbf{Output}: Filter $\mathcal{M}_{Fi}$

    \textcolor[HTML]{36600E}{\textbf{\textit{/ * Feature extraction * /}}}
	
    $\mathcal{M}_G = \mathcal{M}_E \oplus \mathcal{M}_{cl}$
	
	\For{global round $w = 1$ \textbf{to} $W$}{
		Server sends $\mathcal{M}_G^w$ to all clients $C_n$
		
		\For{each client $C_n$}{
			$C_n$ performs $e$ local training rounds:
			 $\theta_{\mathcal{M}_n}^{w,e} = \theta_{\mathcal{M}_n}^w - l_c \cdot \nabla_{\theta} \mathcal{L}(\mathcal{M}_n, \mathcal{D}_n^k)$
		}
		
        $\theta_{\mathcal{M}_G}^{w+1} = \sum_{n=1}^{N} \frac{|\mathcal{D}_n|}{|\mathcal{D}|} \theta_{\mathcal{M}_n}^{w,e}$

	}

	\textcolor[HTML]{36600E}{\textbf{\textit{/ * Generation of class centroid samples * /}}}
    
	\For{each client $C_n$}{
        $V_n^k = \mathcal{M}_E(x_i^k),x_i^k \in \mathcal{D}_n$
		
        ${V'}_n^k = V_n^k \setminus \{ \mathcal{M}_E(x_i^k) | x_i^k \in \mathcal{D}_n^U \}$
		
        $\mu_n^{R,k} = \mathrm{KMeans}(V_n^{R,k}, K_{n,k})$

        Generate $\tilde{\mu}_n^{R,k}$ via Eq.~\eqref{noise}
	}
    
	\textcolor[HTML]{36600E}{\textbf{\textit{/ * One-shot federated unlearning * /}}}
    $\mathcal{M}_G' = \mathcal{M}_E \oplus \mathcal{M}_{Fi} \oplus \mathcal{M}_{cl}$
	
	Freeze parameters of 
    $\mathcal{M}_\mathcal{G}$: $\theta_{\mathcal{M}_\mathcal{G}} = \theta_{\mathcal{M}_\mathcal{G}}^{*}$

	Train filter $\mathcal{M}_{Fi}$ using $\tilde{\mu}_\mathcal{G}^{R,k}$ via Eq.~\eqref{filter loss} and Eq.~\eqref{update parameters}

	\textcolor[HTML]{36600E}{\textbf{\textit{/ * Restoration * /}}}

    Remove filter: $\mathcal{M}_{Fi}$: $\mathcal{M}_G = \mathcal{M}_E \oplus \mathcal{M}_{cl}$\
\end{algorithm}

 The algorithm, as illustrated in Algorithm \ref{alg:one-shot}, consists of three stages: feature extraction, generation of class centroid samples $\mu_n^{R,k}$ and one-shot federated unlearning. Upon receiving an unlearning request, each client first removes the corresponding data points from its local feature set and computes class centroid samples using the retained data. These centroids are then protected by a differential privacy mechanism and uploaded to the server. The server aggregates the uploaded centroids from all clients to obtain global retained class centroids as $\tilde{\mu}_\mathcal{G}^{R,k}$. Then the server freezes the parameters of the original global model $\mathcal{M}_\mathcal{G}$ and randomly initializes a lightweight, pluggable filter $\mathcal{M}_{Fi}$, which is trained solely using the global differentially private centroids. By jointly optimizing a cross-entropy loss and a reconstruction loss $\mathcal{L}_{\text{total}}$, the filter blocks the propagation of forgotten knowledge while preserving the discriminative capability of retained knowledge. Finally, when restoration of forgotten knowledge is required, the filter can be removed to revert to the original global model structure, enabling efficient and reversible federated unlearning.

\section{Experiment}
\subsection{Experimental Setup}

We conducted experiments on diverse datasets, including MNIST\cite{lecun2002gradient}, CIFAR-10, CIFAR-100 \cite{krizhevsky2009learning}, and AG News\cite{zhang2015character}. We partitioned the dataset using the Dirichlet distribution \cite{li2022federated} with a concentration parameter of 0.5. These experiments use various network architectures, including LeNet-5, ResNet-18 \cite{he2016deep}, ResNet-34, Transformer \cite{vaswani2017attention} and TinyBert\cite{jiao2020tinybert}, covering three unlearning scenarios: client unlearning, class unlearning and sample unlearning.
We evaluated five baselines: EraseClient \cite{Halimi_Kadhe_Rawat_Baracaldo_2022}, Federaser \cite{liu2021federaser}, Exact-Fun \cite{xiong2023exact}, Retrain and FUSED \cite{zhong2025unlearning}. 
The experimental framework is implemented using PyTorch 2.3.1 and CUDA 12.1. For hardware acceleration, an NVIDIA RTX 3080 Ti GPU is utilized. We employ SGD and Adam optimizers. The balancing factor $\alpha$ for the loss of filters is set to 0.5. During the generation of sample centroids via clustering, we set different sampling ratios $\rho$ to accommodate various scenarios. In the client, class, and sample scenarios, the sampling ratios are 0.8, 0.1, and 0.5, respectively.

\begin{table}[h]
\centering
\fontsize{8pt}{9pt}\selectfont
\setlength{\tabcolsep}{8pt}
\setlength{\aboverulesep}{0pt}
\setlength{\belowrulesep}{0pt}
\setlength{\arrayrulewidth}{1pt} 
\renewcommand{\arraystretch}{1.2} 

\begin{tabular}{>{\centering\arraybackslash}m{1.5cm}|>{\centering\arraybackslash}m{5.5cm}}
\hline
\rowcolor{gray!20}
\textbf{\textit{Metric}} & \textbf{\textit{Description}} \\
\hline
\textit{\textbf{R-A(\%)$\uparrow$}} & Accuracy of retained knowledge. \\
\textit{\textbf{F-A(\%)$\downarrow$}} & Accuracy of unlearning knowledge. \\
\textit{\textbf{0A(\%)$\uparrow$}} & Accuracy of class 0. \\
\textit{\textbf{PS(\%)$\uparrow$}} & Prediction precision of class 0. \\
\textit{\textbf{MIA(\%)$\downarrow$}} & Post-unlearning privacy risk quantified via membership inference attacks. \\
\textit{\textbf{Time(s)$\downarrow$}} & Time required to achieve the desired effect. \\
\textit{\textbf{Comm(MB)$\downarrow$}} & Communication resource consumption between client and server. \\
\hline
\end{tabular}
\caption{Evaluation metrics of FU.}
\label{tab:evaluation-metrics}
\end{table}

\begin{table*}[!t]
\centering
\fontsize{8pt}{9pt}\selectfont
\setlength{\tabcolsep}{0.25pt}
\setlength{\aboverulesep}{0pt}
\setlength{\belowrulesep}{0pt}
\setlength{\arrayrulewidth}{1pt} 
\renewcommand{\arraystretch}{1.2}

\newcolumntype{C}{>{\centering\arraybackslash}p{1.175cm}}
\newcolumntype{L}{>{\centering\arraybackslash\columncolor{gray!20}\bfseries\itshape}p{1.9cm}} 
\newcolumntype{P}{>{\centering\arraybackslash\bfseries\itshape\columncolor{white}}p{0.7cm}} 
\newcolumntype{G}{>{\centering\arraybackslash\columncolor[HTML]{D0E8E0}}p{1.175cm}} 
\newcolumntype{B}{>{\centering\arraybackslash\columncolor[HTML]{E9F1FA}}p{1.175cm}}

\newcolumntype{R}{>{\centering\arraybackslash\columncolor{white}}p{1.175cm}}
\newcolumntype{C}{>{\centering\arraybackslash\columncolor{white}}p{1.175cm}}

\begin{tabularx}{\textwidth}{
  L !{\vrule width 1pt}
  P 
  R C R C B G !{\vrule width 1pt}
  C B G !{\vrule width 1pt}
  P R B G 
}
\hline
\rowcolor{gray!20}
 &  & \multicolumn{6}{c!{\vrule width 1pt}}{\textbf{\textit{Client Unlearning}}} & \multicolumn{3}{c!{\vrule width 1pt}}{\textbf{\textit{Class Unlearning}}} & \multicolumn{4}{c}{\textbf{\textit{Sample Unlearning}}} \\
\rowcolor{gray!20}
\multirow{-2}{*}
{\diagbox[width=1.9cm,font=\bfseries\itshape]{Dataset}{Scenarios}}

&  & \textbf{E-C} & \textbf{Federaser} & \textbf{E-F} & \textbf{FUSED} & \textbf{Retrain} & \textbf{FedUP} & \textbf{FUSED} & \textbf{Retrain} & \textbf{FedUP} &  & \textbf{FUSED} & \textbf{Retrain} & \textbf{FedUP} \\
\hline

 & R-A & 0.99 & 0.97 & 0.97 & 0.99 & 0.99 & \textbf{0.99} & 0.97 & 0.99 & \textbf{1.00} & 0A & 0.95 & \textbf{1.00} & 0.95 \\
 & F-A & 0.00 & 0.00 & 0.00 & 0.00 & 0.00 & \textbf{0.00} & 0.00 & 0.00 & \textbf{0.00} & PS & 1.00 & 1.00 & \textbf{1.00} \\
\multirow{-3}{=}{\centering MNIST-\\LeNet5} 
 & MIA & 0.80 & \textbf{0.60} & 0.76 & 0.69 & 0.67 & 0.70 & 0.99 & \textbf{0.97} & 0.99 & MIA & 0.96 & \textbf{0.68} & 0.97 \\
\hline

 & R-A & 0.57 & 0.46 & 0.56 & 0.56 & \textbf{0.58} & 0.57 & 0.71 & \textbf{0.72} & 0.71 & 0A & 0.60 & \textbf{0.73} & 0.70 \\
 & F-A & 0.04 & 0.08 & 0.04 & 0.05 & 0.03 & \textbf{0.03} & 0.00 & 0.00 & \textbf{0.00} & PS & 0.52 & \textbf{0.60} & 0.53 \\
\multirow{-3}{=}{\centering CIFAR10-\\ResNet18}
 & MIA & 0.67 & 0.70 & 0.62 & 0.68 & \textbf{0.57} & 0.63 & 0.81 & \textbf{0.63} & 0.78 & MIA & \textbf{0.95} & 0.98 & 0.96 \\
\hline

 & R-A & 0.48 & 0.51 & 0.50 & 0.49 & 0.52 & \textbf{0.52} & 0.59 & \textbf{0.60} & 0.59 & 0A & 0.35 & \textbf{0.64} & 0.57 \\
 & F-A & 0.05 & 0.05 & 0.05 & 0.05 & 0.04 & \textbf{0.04} & 0.00 & 0.00 & \textbf{0.00} & PS & 0.50 & \textbf{0.63} & 0.52 \\
\multirow{-3}{=}{\centering CIFAR10-\\Transformer}
 & MIA & 0.54 & \textbf{0.35} & 0.59 & 0.56 & 0.64 & 0.52 & 0.81 & \textbf{0.78} & 0.90 & MIA & 0.96 & 0.91 & \textbf{0.76} \\
\hline

 & R-A & 0.89 & 0.88 & 0.88 & 0.89 & 0.89 & \textbf{0.89} & 0.89 & 0.93 & \textbf{0.93} & 0A & 0.92 & 0.93 & \textbf{0.93} \\
 & F-A & \textbf{0.03} & 0.05 & 0.04 & 0.04 & 0.03 & 0.05 & 0.00 & 0.00 & \textbf{0.00} & PS & 0.87 & \textbf{0.89} & 0.88 \\
\multirow{-3}{=}{\centering AG News-\\TinyBert}
 & MIA & \textbf{0.63} & 0.63 & 0.79 & 0.68 & 0.68 & 0.77 & \textbf{0.44} & 0.63 & 0.50 & MIA & 0.77 & \textbf{0.54} & 0.70 \\
\hline

 & R-A & 0.13 & 0.15 & 0.27 & 0.26 & \textbf{0.27} & 0.26 & 0.36 & 0.37 & \textbf{0.37} & 0A & 0.50 & \textbf{0.57} & 0.55 \\
 & F-A & 0.01 & 0.01 & 0.01 & 0.01 & \textbf{0.01} & 0.02 & 0.00 & 0.00 & \textbf{0.00} & PS & 0.55 & \textbf{0.56} & 0.54 \\
\multirow{-3}{=}{\centering CIFAR100-\\ResNet34}
 & MIA & 0.32 & 0.49 & 0.43 & 0.40 & \textbf{0.26} & 0.69 & 0.91 & \textbf{0.38} & 0.67 & MIA & 0.98 & \textbf{0.98} & 0.99 \\
\hline
\end{tabularx}
\caption{Main results. Our method achieves the best or near-best R‑A in most settings while maintaining low F‑A, and it performs particularly well in the class unlearning scenario, indicating that it maximizes model utility while ensuring effective unlearning.}
\label{tab:results}
\end{table*}

\textbf{Evaluation Metrics}. As shown in Table \ref{tab:evaluation-metrics}, we evaluate our proposed method and the baselines using multiple metrics.

\subsection{Experimental Results}\label{Experimental Results}
\textbf{Main Results}. In the client unlearning setting, Byzantine attacks are introduced, where label flipping is applied to construct inverse feature prototypes \cite{fu2025federated,qi2023cross}. 
Class unlearning randomly remaps the feature labels of the target class to several classes.
Sample unlearning is implemented via backdoor attacks: fixed triggers are injected into the bottom-right pixels of images or appended as specific trigger tokens to the end of text sequences. During the unlearning process, all triggered samples are consistently predicted as class 0, thereby yielding accuracy of class 0 (0A) on the forgotten samples.

As shown in Table \ref{tab:results}, our method achieves superior performance across a wide range of datasets, model architectures, and unlearning scenarios. Among them, CIFAR-100 is the most challenging benchmark, while MNIST is the least. Benefiting from the pluggable filters, FedUP reduces the accuracy on the unlearning knowledge to a minimal level (F-A) while maintaining high accuracy on the retained data (R-A). Owing to its centroid-based unlearning mechanism, FedUP exhibits particularly strong performance in class unlearning scenarios. Overall, the proposed method supports single-round communication and reversible unlearning, achieving the desirable triad of high retention accuracy, low forgetting accuracy, and privacy guarantees across all evaluated datasets.

\begin{figure}[!h]
	\centering
	\includegraphics[width=0.48\textwidth]{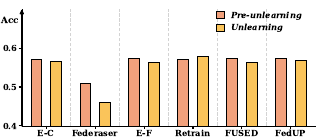}
	\caption{Non-target knowledge loss of methods.}
	\label{non_target_knowledge_loss}
\end{figure}

\textbf{Analysis of Non-target Knowledge Loss}.
To validate the effectiveness of our method in mitigating non-target knowledge loss, we compare the accuracy on retained knowledge before and after executing the unlearning operation. As shown in Figure \ref{non_target_knowledge_loss}, FedUP achieves an effect comparable to the Retrain method, characterized by minimal non-target knowledge loss and stable model performance before and after the unlearning operation.

\textbf{Unlearning Response Time}. 
As shown in Table \ref{tab:Unlearning Cost}, when achieving the desired forgetting effect, our method requires only 6.33 seconds for image unlearning and 6.45 seconds for text unlearning, which is less than 10\% of the latency of the second-best method and less than 1\% of that of the Retrain approach. It can promptly respond to diverse unlearning requests across different modalities while maintaining consistently low latency, effectively minimizing processing delays.

\textbf{Communication Cost}. Communication cost is defined as the wireless resources consumed for transmitting model parameters or gradients between local clients and the central server. As shown in Table \ref{tab:Unlearning Cost}, although the centroid uploading phase introduces additional communication overhead, FedUP requires only a single communication round, meaning its communication cost does not accumulate with rounds. In contrast, model transmission methods gradually converge as rounds increase, causing the communication overhead to escalate to the order of $10^3$. Specifically, the lightweight filter incurs an overhead of merely 11.50MB for image unlearning, approximately one-third of that of the second-best method and 6.73MB for text unlearning, which is second only to the FUSED method.

\begin{table}[h]
\centering
\fontsize{8pt}{9pt}\selectfont
\setlength{\tabcolsep}{9.3pt}
\setlength{\aboverulesep}{0pt}
\setlength{\belowrulesep}{0pt}
\setlength{\arrayrulewidth}{1pt} 
\renewcommand{\arraystretch}{1.2}

\newcolumntype{L}{>{\centering\arraybackslash}p{1.85cm}}

\begin{tabular}{L|cc|cc}
\hline
\rowcolor{gray!20}
 & \multicolumn{2}{c|}{\textbf{\textit{CIFAR10}}} & \multicolumn{2}{c}{\textbf{\textit{AG News}}} \\
\rowcolor{gray!20}
\multirow{-2}{*}{\diagbox[width=2.35cm,font=\bfseries\itshape]{Dataset}{Methods}}
 & \textbf{Time} & \textbf{Comm} & \textbf{Time} & \textbf{Comm} \\
\hline
\rowcolor{white}
\textbf{\textit{E-C}}       & 1627.76 & 1110.98 & 567.56  & 268.64 \\
\rowcolor{white}
\textbf{\textit{Federaser}} & 1586.59 & 2820.18 & 634.67  & 738.76 \\
\rowcolor{white}
\textbf{\textit{E-F}}       & 898.70  & 1452.82 & 629.83  & 369.38 \\
\rowcolor{white}
\textbf{\textit{FUSED}}     & 151.21  & 31.36  & 88.31   & \textbf{0.55}  \\
\rowcolor[HTML]{E9F1FA}
\textbf{\textit{Retrain}}   & 935.54 & 1538.25 & 859.28  & 537.28 \\
\rowcolor[HTML]{D0E8E0}
\textbf{\textit{FedUP}}       & \textbf{6.33} & \textbf{11.50} & \textbf{6.45} & 6.73 \\
\hline
\end{tabular}
\caption{Comparing time and communication costs.}
\label{tab:Unlearning Cost}
\end{table}

\begin{figure*}[!ht]
	\centering
	\includegraphics[width=0.99\textwidth]{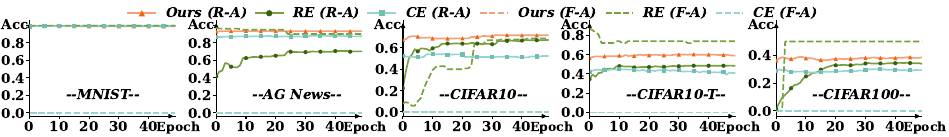}
	\caption{Ablation of loss functions.}
	\label{fig:ablation_loss}
\end{figure*}

\subsection{Analysis of Hyper-parameters}
\textbf{Bottleneck Dimension}. In this section, we analyse the dimension of the filter. The pluggable filter adopts a straightforward encoder-decoder architecture. Its input dimension is configured to match the output dimension of the feature extractor. Our empirical investigation focuses on determining the optimal bottleneck dimension within this encoder-decoder structure. As shown in Table \ref{tab:adapter_dim}, the remember accuracy consistently reaches its optimum across both image and text datasets when the bottleneck dimension is set to 32.

\begin{table}[!ht]
    \centering
    \fontsize{8pt}{9pt}\selectfont
    \setlength{\tabcolsep}{0.5pt}
    \setlength{\aboverulesep}{0pt}
    \setlength{\belowrulesep}{0pt}
    \setlength{\arrayrulewidth}{1pt} 
    \renewcommand{\arraystretch}{1.2}
    
    \newcolumntype{C}{>{\centering\arraybackslash}p{0.60cm}} 
    \newcolumntype{L}{>{\centering\arraybackslash}p{1.99cm}} 

    \begin{tabularx}{0.48\textwidth}{L| CC | CC | CC | CC | CC}
        \hline
        \rowcolor{gray!20}
        & \multicolumn{2}{c|}{\textbf{\textit{4}}} 
        & \multicolumn{2}{c|}{\textbf{\textit{8}}} 
        & \multicolumn{2}{c|}{\textbf{\textit{16}}} 
        & \multicolumn{2}{c|}{\textbf{\textit{32}}} 
        & \multicolumn{2}{c}{\textbf{\textit{64}}} \\
        
        \rowcolor{gray!20}
        \multirow{-2}{*}{\diagbox[width=1.99cm+2\tabcolsep,font=\fontsize{9pt}{10pt}\selectfont\bfseries\itshape]{Dataset}{Dimension}}
        & \textbf{R-A} & \textbf{M} & \textbf{R-A} & \textbf{M} & \textbf{R-A} & \textbf{M} & \textbf{R-A} & \textbf{M} & \textbf{R-A} & \textbf{M} \\
        \hline
        \rowcolor{white}
        \textbf{\textit{MNIST}}     & 0.98 & 16   & 0.99 & 32   & 0.99 & 64   & \textbf{0.99} & 128  & 0.99 & 256 \\
        \rowcolor{white}
        \textbf{\textit{CIFAR10}}   & 0.55 & 16   & 0.58 & 32   & 0.58 & 64   & \textbf{0.60} & 128  & 0.59 & 256 \\
        \rowcolor{white}
        \textbf{\textit{CIFAR10-T}} & 0.43 & 16   & 0.47 & 32   & 0.46 & 64   & \textbf{0.51} & 128  & 0.46 & 256 \\
        \rowcolor{white}
        \textbf{\textit{AG News}}   & 0.89 & 4    & 0.90 & 8    & 0.90 & 16   & \textbf{0.91} & 32   & 0.90 & 64  \\
        \rowcolor{white}
        \textbf{\textit{CIFAR100}}  & 0.06 & 16   & 0.18 & 32   & 0.24 & 64   & \textbf{0.25} & 128  & 0.23 & 256 \\
        
        \hline
    \end{tabularx}
    \caption{Exploration of filter. ``M" stands for memory (KB).}
    \label{tab:adapter_dim}
\end{table}

\textbf{Sensitivity Analysis of the Loss Function.} The filter is trained with joint reconstruction and cross-entropy losses. We performed a sensitivity analysis on $\alpha$ of $\mathcal{L}_{\text{total}}$ to determine its optimal value.
 As shown in Figure \ref{fig:ablation_loss}, using either loss alone hinders convergence and degrades the accuracy of retained knowledge, whereas their combination maximizes the reduction of non-target knowledge loss. This is further illustrated in Table \ref{tab:alpha}, which shows that increasing the weight of the cross-entropy loss lowers the recognition accuracy on retained data, while assigning a higher weight to the reconstruction loss deteriorates the unlearning effectiveness.

\begin{table}[!ht]
    \centering
    \fontsize{8pt}{9pt}\selectfont
    \setlength{\tabcolsep}{0.5pt}
    \setlength{\aboverulesep}{0pt}
    \setlength{\belowrulesep}{0pt}
    \setlength{\arrayrulewidth}{1pt} 
    \renewcommand{\arraystretch}{1.2}
    
    \newcolumntype{C}{>{\centering\arraybackslash}p{0.647cm}} 
    \newcolumntype{L}{>{\centering\arraybackslash}p{1.5cm}} 

    \begin{tabularx}{0.48\textwidth}{L| CC | CC | CC | CC | CC}
        \hline
        \rowcolor{gray!20}
        \textbf{\textbf{\textit{$\alpha$-CE}}}
         & \multicolumn{2}{c|}{\textbf{\textit{0.10}}}
         & \multicolumn{2}{c|}{\textbf{\textit{0.30}}}
         & \multicolumn{2}{c|}{\textbf{\textit{0.50}}}
         & \multicolumn{2}{c|}{\textbf{\textit{0.70}}}
         & \multicolumn{2}{c}{\textbf{\textit{0.90}}} \\
        
        \rowcolor{gray!20}
        \textbf{metric}
         & \textbf{R-A} & \textbf{F-A}
         & \textbf{R-A} & \textbf{F-A}
         & \textbf{R-A} & \textbf{F-A}
         & \textbf{R-A} & \textbf{F-A}
         & \textbf{R-A} & \textbf{F-A} \\
        \hline
        \rowcolor{white}
        \textbf{\textit{MNIST}}
         & 0.99 & 0.67
         & \textbf{1.00} & 0.00
         & 0.99 & 0.00
         & 0.99 & 0.00
         & 0.99 & 0.00 \\
        \rowcolor{white}
        \textbf{\textit{CIFAR10}}
         & 0.70 & 0.00
         & 0.68 & 0.00
         & \textbf{0.71} & 0.00
         & 0.68 & 0.00
         & 0.64 & 0.00 \\
        \rowcolor{white}
        \textbf{\textit{CIFAR10-T}}
         & 0.51 & 0.22
         & 0.56 & 0.00
         & \textbf{0.59} & 0.00
         & 0.51 & 0.00
         & 0.45 & 0.00 \\
        \rowcolor{white}
        \textbf{\textit{AG News}}
         & 0.92 & 0.00
         & 0.93 & 0.00
         & \textbf{0.93} & 0.00
         & 0.92 & 0.00
         & 0.87 & 0.00 \\
        \rowcolor{white}
        \textbf{\textit{CIFAR100}}
         & 0.35 & 0.11
         & 0.37 & 0.00
         & \textbf{0.37} & 0.00
         & 0.31 & 0.00
         & 0.30 & 0.00 \\
        
        \hline
    \end{tabularx}
    \caption{Impact of $\alpha$-CE on R-A and F-A.
    The filter shows optimal and well-balenced performance at $\alpha = 0.5$.
    }
    \label{tab:alpha}
\end{table}

\subsection{Ablation Study}

\textbf{Analysis of Differential Privacy Effects}.
To ensure privacy preservation, Gaussian noise is injected into the generated class centroids. We conducted ablation studies to evaluate its impact, as shown in Table \ref{dp}. The results demonstrate that an appropriately calibrated noise scale does not significantly impede the model's performance on non-target knowledge. Instead, it enhances model robustness, achieving a synergistic improvement in both privacy protection and model utility.

\begin{table}[!ht]
\centering
\fontsize{8pt}{9pt}\selectfont
\setlength{\tabcolsep}{6.35pt} 
\setlength{\aboverulesep}{0pt}
\setlength{\belowrulesep}{0pt}
\setlength{\arrayrulewidth}{1pt}
\renewcommand{\arraystretch}{1.2}

\newcolumntype{L}{>{\centering\arraybackslash}p{1.7cm}}  
\newcolumntype{C}{>{\centering\arraybackslash}p{1.1cm}}  

\begin{tabular}{L|CC|CC}
\hline
\rowcolor{gray!20}
 & \multicolumn{2}{c|}{\textbf{\textit{CIFAR10}}} & \multicolumn{2}{c}{\textbf{\textit{AG News}}} \\
\rowcolor{gray!20}
\multirow{-2}{*}{\diagbox[width=2cm,font=\bfseries\itshape]{Metrics}{Dataset}}
 & \textbf{DP} & \textbf{w/o DP} & \textbf{DP} & \textbf{w/o DP} \\
\hline
\textbf{\textit{R-A}}      & 0.5814 & \textbf{0.5833} & \textbf{0.9132} & 0.9114 \\
\textbf{\textit{F-A}}    & \textbf{0.0438} & 0.0460 & 0.0490 & \textbf{0.0303} \\
\textbf{\textit{MIA}}  & \textbf{0.5239} & 0.6373 & \textbf{0.5654} & 0.6196 \\
\hline
\end{tabular}
\caption{Differential privacy effects. 
}
\label{dp}
\end{table}

\begin{table}[!ht]
\centering
\fontsize{8pt}{9pt}\selectfont
\setlength{\tabcolsep}{5.9pt} 
\setlength{\aboverulesep}{0pt}
\setlength{\belowrulesep}{0pt}
\setlength{\arrayrulewidth}{1pt}
\renewcommand{\arraystretch}{1.2}
\newcolumntype{L}{>{\centering\arraybackslash}p{1.7cm}}  
\newcolumntype{C}{>{\centering\arraybackslash}p{1.1cm}} 
\begin{tabular}{L|CC|CC}
\hline
\rowcolor{gray!20}
 & \multicolumn{2}{c|}{\textbf{\textit{R-A}}} & \multicolumn{2}{c}{\textbf{\textit{F-A}}} \\
\rowcolor{gray!20}
\multirow{-2}{*}{\diagbox[width=2cm,font=\bfseries\itshape]{Dataset}{Acc}}
 & \textbf{Feature} & \textbf{Centroid} & \textbf{Feature} & \textbf{Centroid} \\
\hline
\textbf{\textit{MNIST}}      & 0.9900 & \textbf{0.9921} & \textbf{0.0001} & 0.0005 \\
\textbf{\textit{CIFAR10}}    & 0.6011 & \textbf{0.6020} & 0.0498 & \textbf{0.0331} \\
\textbf{\textit{CIFAR10-T}}  & 0.5264 & \textbf{0.5671} & 0.0632 & \textbf{0.0490} \\
\textbf{\textit{AG News}}    & \textbf{0.9153} & 0.9118 & \textbf{0.0331} & 0.0482 \\
\textbf{\textit{CIFAR100}}   & \textbf{0.2537} & 0.2508 & \textbf{0.0076} & 0.0172 \\
\hline
\end{tabular}
\caption{Ablation of class centroid samples.}
\label{tab:ablation_class_centroid}
\end{table}

\textbf{Effectiveness of Class Centroid Samples}. Class centroids are obtained by clustering original features and privately aggregated on the server side. To validate the effectiveness of class centroid samples, we additionally conduct experiments in which the filter is trained using only the original features. As illustrated in Table \ref{tab:ablation_class_centroid}, the model trained on class centroid samples achieves comparable performance on the accuracy of retained and unlearning knowledge to that trained on the original features, demonstrating that class centroid samples are as effective as the original features.

\section{Conclusion and Discussion}
\textbf{Conclusion}.
In the field of FU, server-side methods face non-target knowledge loss, whereas client‑side methods incur high request latency. Both are limited by the irreversibility of the unlearning operation. To address these issues, this paper proposes a one-shot federated unlearning framework. By employing differentially private class centroid samples on the server, our approach achieves approximate unlearning that surpasses exact unlearning in effect, reducing non‑target knowledge loss and high resource overhead. Through fine-tuning a lightweight plug-in filter in a single round, the desired unlearning effect is achieved, significantly reducing the latency of unlearning responses. Removing the filter allows the model to revert to its pre-unlearning state, thus realizing the reversibility of unlearning. Extensive experiments across diverse datasets, scenarios, and models demonstrate that FedUP demonstrates excellent performance.

\textbf{Discussion}.
While FedUP advances unlearning efficiency, reversibility, and responsiveness, two fundamental limitations still persist: Class centroid fidelity depends on federated feature extraction. Low-quality global models propagate bias into aggregated class centroids. It is expected that these limitations can be effectively addressed by utilizing more powerful pre-trained backbones for local feature extraction to ensure reliable class centroids.

\subsubsection{Contribution Statement}
Feihong Nan and Zhengyi Zhong contributed equally.

\bibliographystyle{named}
\bibliography{ijcai26}

@article{de_la_Torre_2018,  
 title={A Guide to the California Consumer Privacy Act of 2018}, 
 url={http://dx.doi.org/10.2139/ssrn.3275571}, 
 DOI={10.2139/ssrn.3275571}, 
 journal={SSRN Electronic Journal}, 
 author={de la Torre, Lydia}, 
 year={2018}, 
 month={Dec}, 
 language={en-US} 
 }

@article{McMahan_Moore_Ramage_Hampson_Arcas_2016,  
 title={Communication-Efficient Learning of Deep Networks from Decentralized Data}, 
 journal={arXiv: Learning,arXiv: Learning}, 
 author={McMahan, H.Brendan and Moore, EiderB and Ramage, Daniel and Hampson, Seth and Arcas, BlaiseAgüeray}, 
 year={2016}, 
 month={Feb}, 
 language={en-US} 
 }

@article{wu2022federated,
  title={Federated unlearning with knowledge distillation},
  author={Wu, Chen and Zhu, Sencun and Mitra, Prasenjit},
  journal={arXiv preprint arXiv:2201.09441},
  year={2022}
}

@article{huynh2025certified,
  title={Certified unlearning for federated recommendation},
  author={Huynh, Thanh Trung and Nguyen, Trong Bang and Nguyen, Thanh Toan and Nguyen, Phi Le and Yin, Hongzhi and Nguyen, Quoc Viet Hung and Nguyen, Thanh Tam},
  journal={ACM Transactions on Information Systems},
  volume={43},
  number={2},
  pages={1--29},
  year={2025},
  publisher={ACM New York, NY}
}

@inproceedings{pan2025federated,
  title={Federated unlearning with gradient descent and conflict mitigation},
  author={Pan, Zibin and Wang, Zhichao and Li, Chi and Zheng, Kaiyan and Wang, Boqi and Tang, Xiaoying and Zhao, Junhua},
  booktitle={Proceedings of the AAAI Conference on Artificial Intelligence},
  volume={39},
  number={19},
  pages={19804--19812},
  year={2025}
}

@inproceedings{wang2023bfu,
  title={Bfu: Bayesian federated unlearning with parameter self-sharing},
  author={Wang, Weiqi and Tian, Zhiyi and Zhang, Chenhan and Liu, An and Yu, Shui},
  booktitle={Proceedings of the 2023 ACM Asia Conference on Computer and Communications Security},
  pages={567--578},
  year={2023}
}

@inproceedings{liu2022right,
  title={The right to be forgotten in federated learning: An efficient realization with rapid retraining},
  author={Liu, Yi and Xu, Lei and Yuan, Xingliang and Wang, Cong and Li, Bo},
  booktitle={IEEE INFOCOM 2022-IEEE conference on computer communications},
  pages={1749--1758},
  year={2022},
  organization={IEEE}
}

@inproceedings{zhu2023heterogeneous,
  title={Heterogeneous federated knowledge graph embedding learning and unlearning},
  author={Zhu, Xiangrong and Li, Guangyao and Hu, Wei},
  booktitle={Proceedings of the ACM web conference 2023},
  pages={2444--2454},
  year={2023}
}

@inproceedings{zhong2025unlearning,
  title={Unlearning through knowledge overwriting: Reversible federated unlearning via selective sparse adapter},
  author={Zhong, Zhengyi and Bao, Weidong and Wang, Ji and Zhang, Shuai and Zhou, Jingxuan and Lyu, Lingjuan and Lim, Wei Yang Bryan},
  booktitle={Proceedings of the Computer Vision and Pattern Recognition Conference},
  pages={30661--30670},
  year={2025}
}

@inproceedings{yang2025erase,
  title={Erase then Rectify: A Training-Free Parameter Editing Approach for Cost-Effective Graph Unlearning},
  author={Yang, Zhe-Rui and Han, Jindong and Wang, Chang-Dong and Liu, Hao},
  booktitle={Proceedings of the AAAI Conference on Artificial Intelligence},
  volume={39},
  number={12},
  pages={13044--13051},
  year={2025}
}

@article{kuo2025exact,
  title={Exact unlearning of finetuning data via model merging at scale},
  author={Kuo, Kevin and Setlur, Amrith and Srinivas, Kartik and Raghunathan, Aditi and Smith, Virginia},
  journal={arXiv preprint arXiv:2504.04626},
  year={2025}
}

@inproceedings{Yinzhi2018Efficient,
title={Efficient Repair of Polluted Machine Learning Systems via Causal Unlearning},
author={Yinzhi Cao and Alexander Fangxiao Yu and Andrew Aday and Eric Stahl and Jon Merwine and Junfeng Yang},
booktitle={ASIACCS '18: Proceedings of the 2018 on Asia Conference on Computer and Communications Security},
year={2018},
}

@article{kurmanji2023towards,
  title={Towards unbounded machine unlearning},
  author={Kurmanji, Meghdad and Triantafillou, Peter and Hayes, Jamie and Triantafillou, Eleni},
  journal={Advances in neural information processing systems},
  volume={36},
  pages={1957--1987},
  year={2023}
}

@inproceedings{golatkar2020eternal,
  title={Eternal sunshine of the spotless net: Selective forgetting in deep networks},
  author={Golatkar, Aditya and Achille, Alessandro and Soatto, Stefano},
  booktitle={Proceedings of the IEEE/CVF conference on computer vision and pattern recognition},
  pages={9304--9312},
  year={2020}
}

@inproceedings{bourtoule2021machine,
  title={Machine unlearning},
  author={Bourtoule, Lucas and Chandrasekaran, Varun and Choquette-Choo, Christopher A and Jia, Hengrui and Travers, Adelin and Zhang, Baiwu and Lie, David and Papernot, Nicolas},
  booktitle={2021 IEEE symposium on security and privacy (SP)},
  pages={141--159},
  year={2021},
  organization={IEEE}
}

@inproceedings{chen2022graph,
  title={Graph unlearning},
  author={Chen, Min and Zhang, Zhikun and Wang, Tianhao and Backes, Michael and Humbert, Mathias and Zhang, Yang},
  booktitle={Proceedings of the 2022 ACM SIGSAC conference on computer and communications security},
  pages={499--513},
  year={2022}
}

@inproceedings{wang2023inductive,
  title={Inductive graph unlearning},
  author={Wang, Cheng-Long and Huai, Mengdi and Wang, Di},
  booktitle={32nd USENIX Security Symposium (USENIX Security 23)},
  pages={3205--3222},
  year={2023}
}

@inproceedings{graves2021amnesiac,
  title={Amnesiac machine learning},
  author={Graves, Laura and Nagisetty, Vineel and Ganesh, Vijay},
  booktitle={Proceedings of the AAAI Conference on Artificial Intelligence},
  volume={35},
  number={13},
  pages={11516--11524},
  year={2021}
}

@article{gupta2021adaptive,
  title={Adaptive machine unlearning},
  author={Gupta, Varun and Jung, Christopher and Neel, Seth and Roth, Aaron and Sharifi-Malvajerdi, Saeed and Waites, Chris},
  journal={Advances in Neural Information Processing Systems},
  volume={34},
  pages={16319--16330},
  year={2021}
}

@article{guo2019certified,
  title={Certified data removal from machine learning models},
  author={Guo, Chuan and Goldstein, Tom and Hannun, Awni and Van Der Maaten, Laurens},
  journal={arXiv preprint arXiv:1911.03030},
  year={2019}
}

@inproceedings{wang2022federated,
  title={Federated unlearning via class-discriminative pruning},
  author={Wang, Junxiao and Guo, Song and Xie, Xin and Qi, Heng},
  booktitle={Proceedings of the ACM web conference 2022},
  pages={622--632},
  year={2022}
}

@inproceedings{liu2021federaser,
  title={Federaser: Enabling efficient client-level data removal from federated learning models},
  author={Liu, Gaoyang and Ma, Xiaoqiang and Yang, Yang and Wang, Chen and Liu, Jiangchuan},
  booktitle={2021 IEEE/ACM 29th International Symposium on Quality of Service (IWQOS)},
  pages={1--10},
  year={2021},
  organization={IEEE}
}

@article{zhong2025sacfl,
  title={Sacfl: Self-adaptive federated continual learning for resource-constrained end devices},
  author={Zhong, Zhengyi and Bao, Weidong and Wang, Ji and Chen, Jianguo and Lyu, Lingjuan and Lim, Wei Yang Bryan},
  journal={IEEE Transactions on Neural Networks and Learning Systems},
  year={2025},
  publisher={IEEE}
}

@article{li2025machine,
  title={Machine unlearning: Taxonomy, metrics, applications, challenges, and prospects},
  author={Li, Na and Zhou, Chunyi and Gao, Yansong and Chen, Hui and Zhang, Zhi and Kuang, Boyu and Fu, Anmin},
  journal={IEEE Transactions on Neural Networks and Learning Systems},
  year={2025},
  publisher={IEEE}
}

@article{krizhevsky2009learning,
  title={Learning multiple layers of features from tiny images},
  author={Krizhevsky, Alex and Hinton, Geoffrey and others},
  year={2009},
  publisher={Toronto, ON, Canada}
}

@inproceedings{he2016deep,
  title={Deep residual learning for image recognition},
  author={He, Kaiming and Zhang, Xiangyu and Ren, Shaoqing and Sun, Jian},
  booktitle={Proceedings of the IEEE conference on computer vision and pattern recognition},
  pages={770--778},
  year={2016}
}

@article{vaswani2017attention,
  title={Attention is all you need},
  author={Vaswani, Ashish and Shazeer, Noam and Parmar, Niki and Uszkoreit, Jakob and Jones, Llion and Gomez, Aidan N and Kaiser, {\L}ukasz and Polosukhin, Illia},
  journal={Advances in neural information processing systems},
  volume={30},
  year={2017}
}

@article{Halimi_Kadhe_Rawat_Baracaldo_2022,  
 title={Federated Unlearning: How to Efficiently Erase a Client in FL?}, 
 author={Halimi, Anisa and Kadhe, Swanand and Rawat, Ambrish and Baracaldo, Nathalie}, 
 year={2022}, 
 month={Jul}, 
 language={en-US} 
 }

@inproceedings{xiong2023exact,
  title={Exact-fun: an exact and efficient federated unlearning approach},
  author={Xiong, Zuobin and Li, Wei and Li, Yingshu and Cai, Zhipeng},
  booktitle={2023 IEEE International Conference on Data Mining (ICDM)},
  pages={1439--1444},
  year={2023},
  organization={IEEE}
}

@article{wei2020federated,
  title={Federated learning with differential privacy: Algorithms and performance analysis},
  author={Wei, Kang and Li, Jun and Ding, Ming and Ma, Chuan and Yang, Howard H and Farokhi, Farhad and Jin, Shi and Quek, Tony QS and Poor, H Vincent},
  journal={IEEE transactions on information forensics and security},
  volume={15},
  pages={3454--3469},
  year={2020},
  publisher={IEEE}
}

@article{truong2021privacy,
  title={Privacy preservation in federated learning: An insightful survey from the GDPR perspective},
  author={Truong, Nguyen and Sun, Kai and Wang, Siyao and Guitton, Florian and Guo, YiKe},
  journal={Computers \& Security},
  volume={110},
  pages={102402},
  year={2021},
  publisher={Elsevier}
}

@article{lecun2002gradient,
  title={Gradient-based learning applied to document recognition},
  author={LeCun, Yann and Bottou, L{\'e}on and Bengio, Yoshua and Haffner, Patrick},
  journal={Proceedings of the IEEE},
  volume={86},
  number={11},
  pages={2278--2324},
  year={2002},
  publisher={Ieee}
}

@article{zhang2015character,
  title={Character-level convolutional networks for text classification},
  author={Zhang, Xiang and Zhao, Junbo and LeCun, Yann},
  journal={Advances in neural information processing systems},
  volume={28},
  year={2015}
}

@inproceedings{jiao2020tinybert,
  title={Tinybert: Distilling bert for natural language understanding},
  author={Jiao, Xiaoqi and Yin, Yichun and Shang, Lifeng and Jiang, Xin and Chen, Xiao and Li, Linlin and Wang, Fang and Liu, Qun},
  booktitle={Findings of the association for computational linguistics: EMNLP 2020},
  pages={4163--4174},
  year={2020}
}

@article{ma2022learn,
  title={Learn to forget: Machine unlearning via neuron masking},
  author={Ma, Zhuo and Liu, Yang and Liu, Ximeng and Liu, Jian and Ma, Jianfeng and Ren, Kui},
  journal={IEEE Transactions on Dependable and Secure Computing},
  volume={20},
  number={4},
  pages={3194--3207},
  year={2022},
  publisher={IEEE}
}

@article{liu2024survey,
  title={A survey on federated unlearning: Challenges, methods, and future directions},
  author={Liu, Ziyao and Jiang, Yu and Shen, Jiyuan and Peng, Minyi and Lam, Kwok-Yan and Yuan, Xingliang and Liu, Xiaoning},
  journal={ACM Computing Surveys},
  volume={57},
  number={1},
  pages={1--38},
  year={2024},
  publisher={ACM New York, NY}
}

@article{nguyen2025survey,
  title={A survey of machine unlearning},
  author={Nguyen, Thanh Tam and Huynh, Thanh Trung and Ren, Zhao and Nguyen, Phi Le and Liew, Alan Wee-Chung and Yin, Hongzhi and Nguyen, Quoc Viet Hung},
  journal={ACM Transactions on Intelligent Systems and Technology},
  volume={16},
  number={5},
  pages={1--46},
  year={2025},
  publisher={ACM New York, NY}
}

@inproceedings{mora2024fedunran,
  title={Fedunran: On-device federated unlearning via random labels},
  author={Mora, Alessio and Dominici, Luca and Bellavista, Paolo},
  booktitle={2024 IEEE International Conference on Big Data (BigData)},
  pages={7955--7960},
  year={2024},
  organization={IEEE}
}

@inproceedings{deng2024enable,
  title={Enable the right to be forgotten with federated client unlearning in medical imaging},
  author={Deng, Zhipeng and Luo, Luyang and Chen, Hao},
  booktitle={International Conference on Medical Image Computing and Computer-Assisted Intervention},
  pages={240--250},
  year={2024},
  organization={Springer}
}

@inproceedings{li2022federated,
  title={Federated learning on non-iid data silos: An experimental study},
  author={Li, Qinbin and Diao, Yiqun and Chen, Quan and He, Bingsheng},
  booktitle={2022 IEEE 38th international conference on data engineering (ICDE)},
  pages={965--978},
  year={2022},
  organization={IEEE}
}

@inproceedings{jiang2026unveiling,
  title={Unveiling and Mitigating Untargeted Poisoning Attacks on Federated Knowledge Graph Embedding},
  author={Jiang, Wenzheng and Liang, Ke and Huang, Wenke and Zhang, Xiongtao and Xu, Zhenxing and Wan, Guancheng and Tan, Cheston and Fan, Flint Xiaofeng and Wang, Ji},
  booktitle={Proceedings of the ACM Web Conference 2026},
  pages={2569--2580},
  year={2026}
}

@article{zhong2022flee,
  title={Flee: A hierarchical federated learning framework for distributed deep neural network over cloud, edge, and end device},
  author={Zhong, Zhengyi and Bao, Weidong and Wang, Ji and Zhu, Xiaomin and Zhang, Xiongtao},
  journal={ACM Transactions on Intelligent Systems and Technology (TIST)},
  volume={13},
  number={5},
  pages={1--24},
  year={2022},
  publisher={ACM New York, NY}
}

@article{qi2024cross,
  title={Cross-training with multi-view knowledge fusion for heterogenous federated learning},
  author={Qi, Zhuang and Meng, Lei and He, Weihao and Zhang, Ruohan and Wang, Yu and Qi, Xin and Meng, Xiangxu},
  journal={arXiv e-prints},
  pages={arXiv--2405},
  year={2024}
}

@article{fu2025learn,
  title={Learn the global prompt in the low-rank tensor space for heterogeneous federated learning},
  author={Fu, Lele and Huang, Sheng and Li, Yuecheng and Chen, Chuan and Zhang, Chuanfu and Zheng, Zibin},
  journal={Neural Networks},
  volume={187},
  pages={107319},
  year={2025},
  publisher={Elsevier}
}

@article{fu2025federated,
  title={Federated domain-independent prototype learning with alignments of representation and parameter spaces for feature shift},
  author={Fu, Lele and Huang, Sheng and Lai, Yanyi and Zhang, Chuanfu and Dai, Hong-Ning and Zheng, Zibin and Chen, Chuan},
  journal={IEEE Transactions on Mobile Computing},
  year={2025},
  publisher={IEEE}
}

@inproceedings{qi2023cross,
  title={Cross-silo prototypical calibration for federated learning with non-iid data},
  author={Qi, Zhuang and Meng, Lei and Chen, Zitan and Hu, Han and Lin, Hui and Meng, Xiangxu},
  booktitle={Proceedings of the 31st ACM international conference on multimedia},
  pages={3099--3107},
  year={2023}
}

\end{document}